%% file: samplepaper.tex
\documentclass[runningheads]{llncs}
\usepackage[T1]{fontenc}
\usepackage{graphicx}
\usepackage{amsmath}
\usepackage{amssymb}
\usepackage{mathrsfs}
\usepackage{tikz}
\usepackage{hyperref}
\usepackage{tabularx}
\usepackage{subcaption}
\usepackage[dvipsnames]{xcolor}
\usepackage[normalem]{ulem}
\usepackage[misc]{ifsym}
\begin{document}
\newcolumntype{L}[1]{>{\raggedright\arraybackslash}p{#1}}
\newcolumntype{C}[1]{>{\centering\arraybackslash}p{#1}}
\newcolumntype{R}[1]{>{\raggedleft\arraybackslash}p{#1}}
\title{Using large language models to generate space-filling optimization problems}
\author{Urban Skvorc\inst{*1}\orcidID{0000-0002-7032-1489} \and Niki van Stein\inst{*2}\orcidID{0000-0002-0013-7969} \and Moritz Seiler\inst{1}\orcidID{ 0000-0002-1750-9060} \and Britta Grimme\inst{1}\orcidID{0009-0008-2282-5130} \and Thomas Bäck\inst{2}\orcidID{0000-0001-6768-1478} \and Heike Trautmann\inst{1,3}\orcidID{0000-0002-9788-8282}}

\institute{Machine Learning and Optimization, Paderborn University, Germany \email{\{urban.skvorc,moritz.seiler,bgrimme,heike.trautmann\}@uni-paderborn.de}\and
Leiden Institute of Advanced Computer Science, Leiden University,  The Netherlands \\
\email{\{n.van.stein,T.H.W.Baeck\}@liacs.leidenuniv.nl}\\ \and
Data Management and Biometrics, University of Twente,  The Netherlands \\
$^*$These authors contributed equally to this work.
}
\authorrunning{Skvorc, van Stein et al.}

\title{LLM Driven Design of Continuous Optimization Problems with Controllable High-level Properties}

\titlerunning{LLM Driven Design of Optimization Problems}
%
%
\maketitle              
\begin{abstract}

Benchmarking in continuous black-box optimisation is hindered by the limited structural diversity of existing test suites such as BBOB. We explore whether large language models embedded in an evolutionary loop can be used to design optimisation problems with clearly defined high-level landscape characteristics. Using the LLaMEA framework, we guide an LLM to generate problem code from natural-language descriptions of target properties, including multimodality, separability, basin-size homogeneity, search-space homogeneity and global–local optima contrast. Inside the loop we score candidates through ELA-based property predictors. We introduce an ELA-space fitness-sharing mechanism that increases population diversity and steers the generator away from redundant landscapes. 
A complementary basin-of-attraction analysis, statistical testing and visual inspection, verifies that many of the generated functions indeed exhibit the intended structural traits. In addition, a $t$-SNE embedding shows that they expand the BBOB instance space rather than forming an unrelated cluster. The resulting library provides a broad, interpretable, and reproducible set of benchmark problems for landscape analysis and downstream tasks such as automated algorithm selection.

\end{abstract}
%
%
%





\section{Introduction}
The performance of optimization algorithms strongly depends on the problems being solved; no single algorithm performs best across all instances. A promising direction to address this challenge is \textit{Automated Algorithm Selection}~(AAS), which trains models to automatically select the best-performing algorithm for a given problem instance. Machine-learning-based AAS approaches have shown strong potential~\cite{kerschke2019AS}, but they are constrained by the limited diversity of existing benchmark problems.  
As highlighted by Cenikj et al.~\cite{cenikj2025landscape}, current benchmark suites often cover only narrow regions of the problem space, hindering generalization and making the evaluation of algorithm selectors highly dependent on how training and testing instances are split.

One way to mitigate this limitation is the automatic generation of optimization problems to populate under-represented regions of the problem space. Several approaches have been proposed, including affine combinations of existing problems~\cite{dietrich2022increasing,vermetten2023ma,vermetten2025ma} and randomly generated tree-like functions~\cite{tian2020recommender,seiler2025randoptgen}. However, affine combinations provide only limited novelty, while random function generators are difficult to control certain high-level landscape properties.

In this paper, we present an initial analysis of using \textit{Large Language Models}~(LLMs) for optimization problem generation. Specifically, we investigate whether an LLM can create problems that are both novel and interpretable by enforcing or verifying high-level properties such as multimodality. In recent years, LLMs have demonstrated strong performance in text generation and \textit{Retrieval-Augmented Generation}~(RAG)~\cite{fan2024surveyragmeetingllms}. Within the field of optimization, LLMs have also been successfully applied to algorithm generation. One example is the \textit{Large Language Model–driven Evolutionary Algorithm} (LLaMEA) framework~\cite{van2025llamea}. By contrast, the use of LLMs for generating optimization problems themselves has received relatively little attention, with the work by Yuhiro et al.~\cite{10.1145/3712255.3726663} being, to the best of our knowledge, the only prior study. 

The goal of this work is to assess whether LLaMEA can also serve as a viable tool for generating interpretable optimization problems. In particular, we study the ability of a LLaMEA-based generator to produce problems with predefined high-level characteristics described in natural language.  We examine five high-level properties~\cite{mersmann2010benchmarking} that have previously been used to characterize benchmark problems: \textit{multimodality, search-space homogeneity, basin-size homogeneity, separability,} and \textit{global-to-local optima contrast}.

The \textbf{contributions} of this paper can be summarized as follows: 
\begin{description}
    \item[First,] we present the first use case of LLM-driven evolutionary computation for the automatic synthesis of optimization problems with pre-specified high-level properties.
    \item[Second,] we extend the LLaMEA framework with an adaptive fitness sharing mechanism that promotes diversity among generated problems.
    \item[Third,] we propose verification techniques to evaluate whether problems exhibit the intended properties and discuss the strengths and limitations of the approach.
    \item[Finally,] we provide an extensive library of verified optimization problems with specific high-level properties, made available as an easy-to-use Python library.
\end{description}

The remainder of this paper is structured as follows: Section~\ref{sec:prior_work} provides a brief overview of prior work, Section~\ref{sec:methodology} describes the proposed methodology, including a  description of the LLaMEA framework and its adaption for problem generation. Section~\ref{sec:preexperiments} reports preliminary experiments conducted to fine-tune the LLaMEA framework prior to problem generation. Section~\ref{sec:results} presents the  experimental verification of whether the generated problems exhibit the desired properties. Section~\ref{sec:conclusion} concludes the paper. All the code for used for the experiments in this paper is available in our repository~\cite{van_stein_2026_18306723}.


\section{Prior work}
\label{sec:prior_work}

Several studies have explored the automatic generation of optimization problems. In the  combinatorial domain, one such example is the work by Ullrich et al~\cite{10.1145/3205651.3208284}). 
In the continuous domain, a lot of research builds upon the well-known BBOB problem set~\cite{hansen:inria-00362633}.  
Muñoz et al.~\cite{9185169} aimed to generate new problems with landscape feature values not yet represented in BBOB, whereas Vermetten et al.~\cite{vermetten2025ma} combined two or more BBOB functions through affine transformations to create new instances. 
Further studies have proposed tree-based problem generators independent of the BBOB suite, such as Tian et al.~\cite{9187549} and Seiler et al.~\cite{seiler2025randoptgen}, while Wang et al.~\cite{wang2025instancegenerationmetablackboxoptimization} applied deep-learning models for this purpose.

The work most closely related to ours is that of Muñoz et al.~\cite{9185169}, who generated new continuous benchmark functions by evolving mathematical function forms whose Exploratory Landscape Analysis (ELA)~\cite{mersmann2011exploratory} features match selected target vectors. Starting from the BBOB set, they computed ELA features to map each instance into a high-dimensional \emph{instance space}, where problems with similar feature values are located close to one another.
They then projected this space into two dimensions to visualize coverage by existing benchmarks and used genetic programming to create new functions whose ELA features matched selected target points, thus filling previously unexplored regions.
This provides precise numeric control of feature combinations but requires expert knowledge to select suitable targets.
In contrast, the LLaMEA approach proposed in this paper offers more interpretable control by linking groups of ELA features to qualitative, so-called high-level, landscape properties (e.g., multimodality, separability, basin‑size homogeneity). The LLM then generates new problem definitions guided by these properties, producing diverse and verifiable landscapes described in human‑readable terms rather than numerical feature goals.

Our generator is based on the 
LLaMEA framework~\cite{van2025llamea}, which integrates evolutionary computation with the generative capabilities of large language models (LLMs). 
LLaMEA enables an automated and interpretable approach to algorithm design. 
Unlike traditional evolutionary algorithms \cite{10.5555/548530}, which evolve numerical representations such as parameter vectors, operator weights, or network topologies, LLaMEA evolves executable algorithmic code. 
Each individual in the population represents a candidate algorithm expressed in Python, originally for solving black-box optimization problems but also applied to combinatorial problems and other real-world tasks \cite{van_Stein_2025,10.1145/3712255.3734288}. 
The role of the LLM is to instantiate and refine these candidates by responding to natural-language prompts that encode evolutionary operators such as mutation and recombination.
Each newly generated (mutated) algorithm is executed and evaluated on a suite of benchmark functions, such as those in the BBOB test-bed, resulting in a quantitative fitness value that guides the subsequent selection phase. 
Related LLM-based systems for automated discovery of meta-heuristics (mostly in the field of combinatorial optimization) include Evolution of Heuristics \cite{liu2024evolution}, ReEvo \cite{ye2024reevo}, and MCTS-AHD \cite{zheng2025monte}, which use evolutionary or tree‑search strategies but focus on generating solvers rather than problems.
In this study, we extend LLaMEA to generate diverse optimization problems with predefined high‑level properties. 
Its modular design and proven performance across problem types make it easily adaptable, and the underlying LLM can be selected according to task requirements. 
In our experiments, \textit{gpt5‑nano} was used, as it provides an effective balance between performance, computational efficiency, and ease of parallelization (see Section~\ref{sec:preexperiments}).


To the best of our knowledge, \textit{LLM-EBG}~\cite{10.1145/3712255.3726663} is the only previously published work using LLMs for automatic generation of optimization benchmarks. 
In LLM-EBG, an evolutionary algorithm uses an LLM to perform crossover and mutation operations on symbolic function representations, while the performance gap between two optimizers—genetic algorithms and differential evolution—serves as the fitness criterion. 
The resulting benchmarks are therefore tailored to highlight performance differences between specific algorithms. 
In contrast, \textit{LLaMEA}, integrates an LLM into an evolutionary framework that optimizes directly in the space of landscape properties rather than algorithmic performance. 
By combining Exploratory Landscape Analysis (ELA) features, property-prediction models, and ELA-based fitness sharing, it generates optimization problems with controllable and verifiable structural characteristics such as multimodality, separability, and basin-size homogeneity.


\section{Methodology}
\label{sec:methodology}
Our goal is to evaluate whether LLaMEA can generate optimization problems that meaningfully complement the 24 BBOB benchmark problems~\cite{bbob2019} and exhibit specific high-level landscape properties. 
Before we expand on the methodology of this work, we introduce some formal definitions. 

Formally, a black-box optimization function can be defined as $f:\mathcal{X}\!\to\!\mathcal{Y}$ where $\mathcal{X} \subseteq \mathbb{R}^d$ is the decision space with dimensionality $d \in \mathbb{N}$ and $\mathcal{Y} \subseteq \mathbb{R}$ is the one-dimensional objective space, as we only consider single-objective continuous optimization problems in this work. 
$\vec x \in \mathcal{X}$ is a known candidate solution and $X = \{\vec x_1, \vec x_2, \ldots, \vec x_n\}$ a set of $n$ such solutions. $y = f(\vec x)$ is the objective value of candidate $\vec x$ and $Y = f(X) = \{f(\vec x) \;|\; \vec x \in X\}$ is the set of objective values. In single-objective optimization, the task of black-box optimization is, w.l.o.g., defined as:
\begin{align}
    \vec x^* = \underset{\vec x \;\in\; \mathcal{X}}{\text{arg\,min}}\; f(\vec x),
\end{align}
where $\vec x^*$ is an optimal solution to instance $f$. In black‑box optimization, the analytical form of $f$ is unknown or impractical to analyze directly. Instead, one relies on black-box optimization algorithms that try to find an approximation of the optimal solution, $\hat{\vec{x}}^*$ s.t. $f(\hat{\vec{x}}^*) - f(\vec x^*) \leq \epsilon$ with $\epsilon > 0$, under a limited evaluation budget.

In studies aiming to characterize black box problems rather than solve them, one relies on \textit{instance features}, since a direct algebraic analysis is not possible.
Instance features are hereby continuous summary statistics that correlate with high-level landscape properties such as a numerical representation of the landscape’s geometry, multimodality, separability, and information content. Formally, the feature space is defined as $\Phi \subseteq \mathbb{R}^m$ where $m \in \mathbb{N}$ is its dimensionality. In this work, we consider \textit{Exploratory Landscape Analysis}~(ELA) features, with their extraction process formally defined as 
\begin{align}
    \text{ELA}: \mathscr{P}_\text{fin}(\mathcal{X} \times \mathcal{Y}) \to \Phi,\quad \phi = \text{ELA}(X, Y)
\end{align}
where $\mathscr{P}_\text{fin}$ denotes the powerset of all finite subsets. The feature vector $\phi \in \Phi$ describes the landscape characteristics of the given black-box instance and depends on sampled candidate solutions $X$ and their fitness values $Y=f(X)$.
Because ELA features depend on the sampling, we compute them multiple times per instance and use their average to reduce stochastic effects:
\begin{align}
    \bar{\phi}_f := \mathbb{E}_{X \sim p_X}\!\left[ \text{ELA}(X, f(X)) \right] \in \Phi.
\end{align}
Here, $p_X$ is some sampling distribution (i.e. uniformly sampling, Latin Hypercube, or Sobol sampling). 

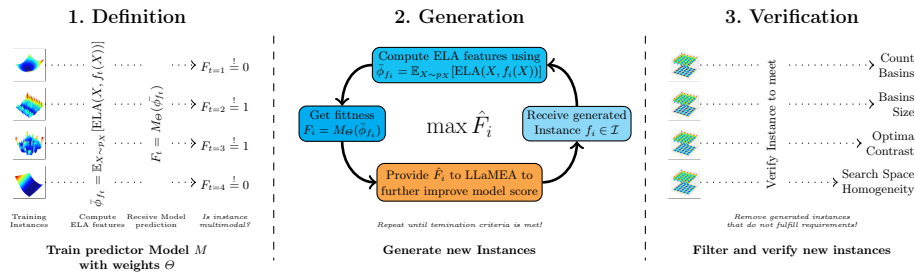
\begin{figure}[t]
    \centering
    \input{figures/abtract_figure}
    \caption{Overview of the three-step LLaMEA process for generating interpretable optimization problems: 
(1) define problem properties and train machine-learning models to predict them from ELA features; 
(2) use the prediction models with an LLM-based generator to create problems exhibiting the properties; 
(3) validate that the generated problems show the desired characteristics. 
}
    \label{fig:abtract}
\end{figure}

The remainder of this section follows the three-step methodology illustrated in Figure~\ref{fig:abtract}: (1.)~Feature model definitions, (2.)~LLM problem generation, and (3.)~Problem feature verification.

\subsection{Feature Model Definition}
We begin our analysis by selecting the high-level problem properties to be generated. The BBOB benchmarking problem set has been well studied in the past, and its problem characteristics are therefore well defined.  Based on the work by Mersmann et al.~\cite{mersmann2010benchmarking}, we focus on the following properties:
\begin{description}
    \item[Multimodality] refers to the number of local optima or basins in the landscape.
    \item[Global–to–Local Optima Contrast] measures the difference between the global and local optima relative to the average fitness. 
    \item[Search–Space Homogeneity] describes the similarity of the landscape structure across the search space.
    \item[Basin–Size Homogeneity] quantifies the variation between the largest and smallest basins.
    \item[Separability] describes the extent to which the landscape can be decomposed into smaller, easier subproblems.
\end{description}


We train machine-learning models on the BBOB problem set to predict whether an optimization problem exhibits specific high-level properties based on its landscape features, following \cite{mersmann2011exploratory,seiler2022collection}.  For each property (except separability, which is assessed directly; see Section~\ref{subsec:separability}), one XGBoost~\cite{chen2016xgboost} regression model is trained using the \texttt{xgboost} Python library with default parameters. 
ELA features are computed with \texttt{pflacco}~\cite{pflacco} from Latin hypercube samples~\cite{mckay2000comparison} of size $250D$ for $D\in\{2,5,10\}$. 
Features are calculated for the first 50 instances of each of the 24 BBOB problems (1\,200 instances per dimensionality).  We use landscape features as opposed to feature-free approaches because landscape features are better interpretable and designed to correlate with high-level properties of the problems we are attempting to generate.
To improve reliability, each problem is sampled five times, and the mean feature values are used. 
We include all features from the groups \texttt{ela\_meta}, \texttt{ela\_distr}, \texttt{nbc}, \texttt{disp}, \texttt{pca}, and \texttt{ic}. 
For each problem, we sample in the domain $x \in (-5,5)$ and rescale fitness values to $y \in (0,1)$ based on the observed minimum and maximum. 
The high-level properties defined in \cite{mersmann2010benchmarking} 
are categorical rather than binary. 
We convert each property into a binary label by grouping the lowest category level versus all higher levels. Binarizing the labels is done to simplify both the training of the model as well as the optimization process of LLaMEA, and to make the results easier to interpret by assigning to each problem a probability of containing the desired property. The resulting models output a score in $[0,1]$ indicating how likely a problem exhibits the corresponding property.

\subsubsection{Evaluating Separability}
\label{subsec:separability}

As noted above, the \emph{separability} property is not well suited for evaluation using a machine learning model, due to the limited number of separable problems in the BBOB set. 
Nevertheless, \emph{separability} remains an important property worth analysing. We therefore adopt the method described in Appendix~A.1.
To summarize, we assess whether an objective function is (approximately) separable 
by numerically testing for variable interactions. 
Specifically, we evaluate the function at randomly sampled points and use finite-difference estimates to measure (1.) how strongly pairs of variables jointly influence the function (via second-order cross terms) and (2.) whether changing one variable alters the marginal effect of changing another (a first-order superposition check). 
Both tests yield violation rates that quantify how often we observe non-negligible interactions. 
We combine these into a single score $p \in [0,1]$, where functions close to $1.0$ are highly likely to be separable.



\subsection{Problem Generation Using LLaMEA}
\label{subsec:problem_generation}
After training the prediction models, the LLaMEA framework is used to generate optimization problems with predefined high-level properties. To improve the diversity of generated optimization landscapes, we extend the original LLaMEA framework \cite{van2025llamea} by integrating a fitness sharing mechanism that operates in the space of ELA features. 
In this setup, each generated function is represented not only by its code and fitness score but also by a vector of ELA features.
By computing pairwise distances between $z$-standardized ELA vectors, we obtain a behaviour-space measure that enables niching in the population of landscapes. 
The following two sections describe the generation process in more detail.

\paragraph{ELA distance metric.} 
Each candidate function is evaluated across several dimensions, \(d \in \{2, 5, 10\}\), using the ELA feature extractors. The resulting features are preprocessed to remove unstable or scale-dependent components and normalized through a pre-trained scaler. The ELA-based distance between two functions \(f_i\) and \(f_j\) is then defined as the Manhattan  distance between their $z$-standardized feature vectors:
\[
D_{\mathrm{ELA}}(f_i, f_j) = || \bar{\phi}_{f_i} - \bar{\phi}_{f_j} ||_1,
\]

\paragraph{Fitness sharing in ELA space.}
The niching strategy in LLaMEA is applied in each generation before selection. Each individual’s fitness is adjusted according to its density in the ELA feature space. The shared fitness \(\hat{F}_i\) of an individual \(i\) is computed as:
\[
\hat{F}_i = \frac{F_i}{\sum_{j \neq i} \max\!\left(0, 1 - \frac{D_{\mathrm{ELA}}(f_i, f_j)}{\sigma_{\text{share}}}\right)},
\]
where \(F_i\) is the raw fitness score obtained from the feature classifiers (e.g., separability, multimodality), and \(\sigma_{\text{share}}\) is the niche radius. This adjustment penalizes individuals who are too similar in ELA space, thereby encouraging exploration of diverse landscape geometries.

\paragraph{Adaptive niche radius.}
To maintain robustness across varying populations and landscape scales, the niche radius \(\sigma_{\text{share}}\) adapts dynamically to the mean pairwise distance in the population in each iteration. This ensures that the sharing mechanism remains effective regardless of the absolute spread of ELA features across generations.

\paragraph{Evaluation and feedback loop.}
The evaluation function integrates multiple diagnostic tests, including separability estimation via finite-difference Hessian cross-terms and superposition checks. Each generated landscape receives quantitative feedback for every targeted feature (e.g., \textit{Separable}, \textit{Global–Local}, \textit{Multimodality}), which is communicated back to the LLM to guide subsequent refinements. Mutation prompts explicitly reference these feature descriptions, allowing the model to iteratively shape landscapes toward/away from specific structural traits.

In the experiments that follow, we used a parent population $\mu = 8$ and an offspring population size of $\lambda = 16$ with a comma strategy (no elitism) to promote exploration, as larger populations tend to provide more exploration as observed in \cite{van2025behaviour}. We generate 55 problems for each group of the following categories (1)~Single-property problems, one for each problem, (2)~Paired-property problems, one for each pair of problems, and (3)~NOT-property problem, problems optimizing the absence of either basin-size homogeneity or search-space homogeneity paired with the presence of 1 other property. 

\subsection{Property Verification}
\begin{figure}[tbp]
  \centering
  \includegraphics[width=0.325\textwidth]{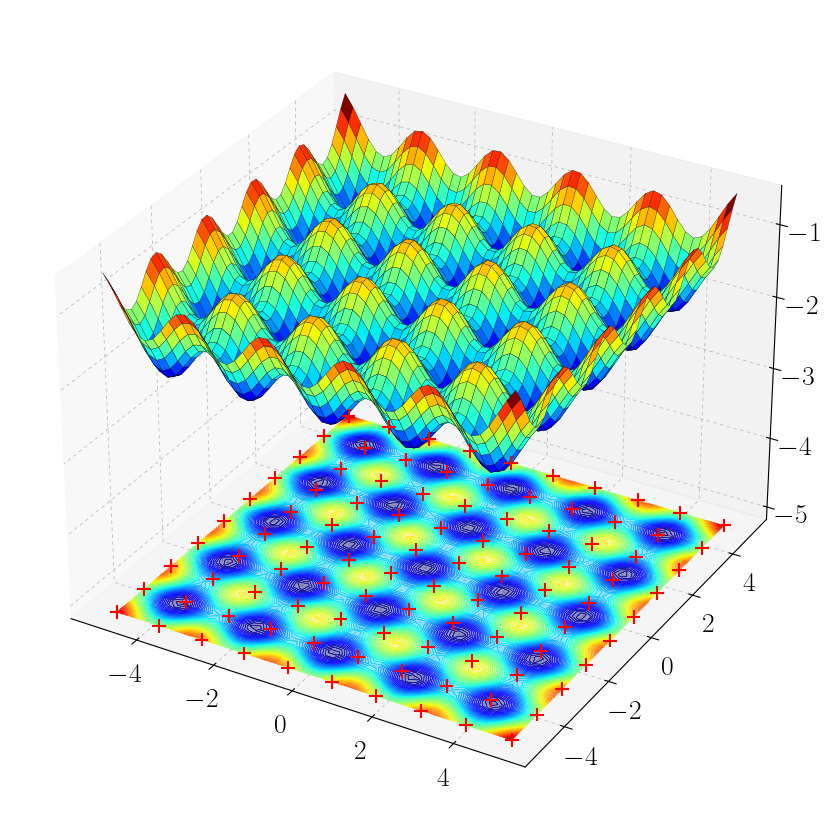}
  \hspace{10mm}
  \includegraphics[width=0.325\textwidth]{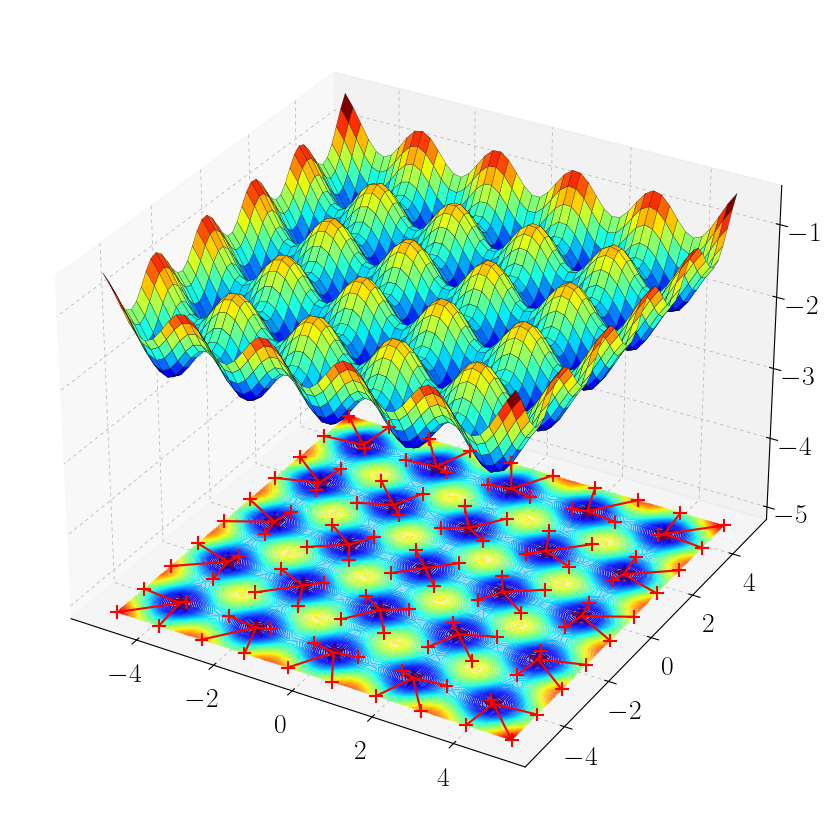}

  \caption{Finding different basins of attraction and their approximate sizes using a dense grid on a $2d$ problem landscape. Each point is attributed to an attractor by iteratively searching for an improving path.}
  \label{fig:basins}
\end{figure}
After problem generation, we need to verify that the generated problem instances truly meet the required high-level properties. The trained property prediction models are fast but do not guarantee fulfilment. Hence, we rely on a different, more costly approach to verify that the generated instance fulfils the defined targets.
We perform an additional property verification using an approach that does not rely on landscape features, but rather on the methodology for analysing basins of attractions introduced by Antonov et al.~\cite{antonov2023new}. This approach also enhances the explainability of our approach by providing simple but informative metrics for each of the analysed properties. While this alternate approach could also be used to train the problems themselves, it is much more computationally expensive than using the property prediction models, especially in higher dimensions. As a result, the analysis is performed only on two dimensional problems. The used method identifies whether two points in the search space belong to the same basin of attraction by iteratively searching for an \emph{improving path}, meaning a continuous sequence of points along which the objective function value does not increase (see Figure~\ref{fig:basins}).

\section{Pre-Experiments}
\label{sec:preexperiments}
Before conducting the main benchmark generation study, we performed two preliminary experiments to guide the design choices for the LLaMEA framework. 
The first experiment investigates the selection of a suitable large language model, while the second evaluates the effectiveness of the proposed ELA-based fitness-sharing mechanism.
\subsection{Selection of the Large Language Model}
\begin{figure}[tbp]
  \centering
  \includegraphics[width=0.375\textwidth, trim={8 20 5 5}, clip]{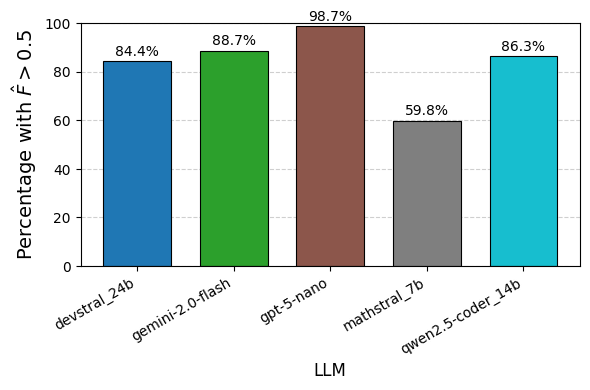}
  \hspace{10mm}
  \includegraphics[width=0.375\textwidth, trim={8 20 5 5}, clip]{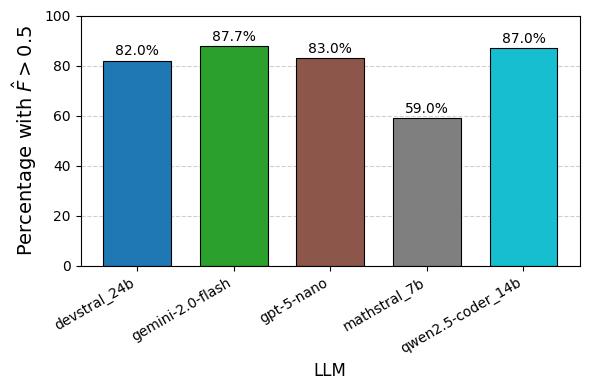}

  \caption{Performance of different LLMs in generating landscapes with certain high‑level properties. Left: results for landscapes optimized for basin‑size homogeneity and separability; Right: multimodal landscapes with a global structure.}
  \label{fig:multiplellms}
\end{figure}

Figure \ref{fig:multiplellms} shows the performance of different LLMs.

To identify a suitable large language model for the LLaMEA framework, we compared five candidates that differ in size and intended focus: 
(1)~\texttt{devstral\_24b}, (2)~\texttt{gemini‑2.0‑flash}, (3)~\texttt{gpt5‑nano}, (4)~\texttt{qwen2.5-coder\_14b}, and (5) \texttt{math-} \texttt{stral\_7b}. 
Each model was integrated into the same LLaMEA optimization pipeline and executed independently to generate problems targeting pairs of high‑level landscape properties. 
During each generation cycle, the property‑pre\-diction models evaluated the current candidate, yielding a fitness score $\hat F$ 
that reflected how strongly the function exhibited the desired characteristics.
When two properties were optimized simultaneously, $\hat F$ was computed as the mean of the two individual property‑prediction scores. 
Figure~\ref{fig:multiplellms} summarizes the proportion of all cycles in which this score exceeded $\hat F>0.5$ (meaning that at least one high-level feature is predicted to be satisfied). 
The left panel corresponds to runs aimed at producing problems that are both \emph{basin‑size homogeneous} and \emph{separable}, whereas the right panel concerns the joint targets of \emph{multimodality} and \emph{global structure}. 

Among the tested models, \texttt{gpt5‑nano} achieved the highest success rate of 98.7\% for the property combination of \emph{basin‑size homogeneity} and \emph{separability}, 
For the second target combination, \emph{multimodality} and \emph{global structure}, the model also ranked among the four best-performing LLMs, all exceeding 80\%. 
Overall, \texttt{gpt5‑nano} was selected for subsequent experiments because it provided the best balance between performance, computational cost, and efficient parallelization.

\subsection{Evaluation of the Fitness Sharing Mechanism}

\begin{figure}[tbp]
  \centering
    \includegraphics[width=0.40\textwidth, trim={20 20 5 5}, clip]{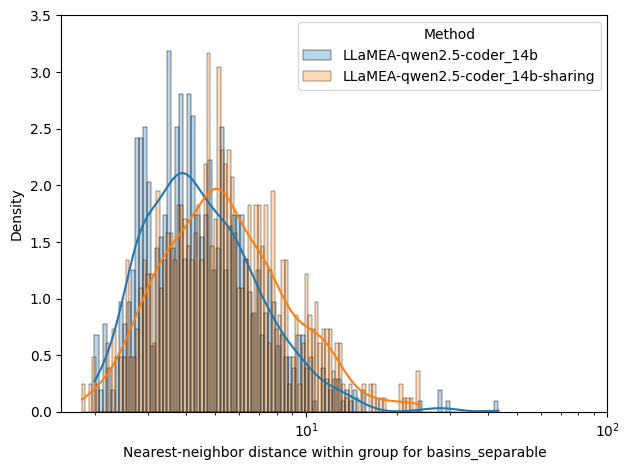}
    \hspace{10mm}
    \includegraphics[width=0.40\textwidth, trim={20 20 5 5}, clip]{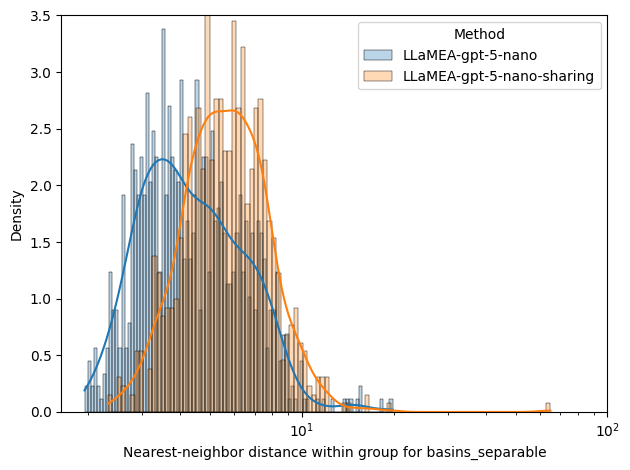}
  \caption{Distribution of nearest-neighbor (Manhattan) distances between normalized ELA feature vectors of landscapes generated with and without fitness sharing. 
Results are shown for \texttt{qwen2.5-coder\_14b} (left) and \texttt{gpt5-nano} (right); larger distances indicate greater landscape diversity. 
} 
  \label{fig:pairwise}
\end{figure}

We first verified whether the proposed ELA‑based fitness‑sharing mechanism indeed increases the diversity of the generated problem landscapes in terms of their ELA features. 
To this end, we compared two LLMs, \texttt{qwen2.5‑coder\_14b}, as it was the best performing open-source model in our initial experiments, and \texttt{gpt5‑nano}, the best closed-source model, each executed both with and without fitness sharing. 
For every run, we computed pairwise Manhattan distances 
between the normalized ELA feature vectors of all generated landscapes. Figure~\ref{fig:pairwise} shows the distributions of nearest‑neighbor distances between the landscapes generated while optimizing for \emph{basin‑size homogeneity} and \emph{separability}.

The distributions obtained with fitness sharing are clearly shifted toward larger distances, indicating that the generated functions became more diverse as intended. 
This effect is particularly pronounced for the \texttt{gpt5‑nano} model, whose peak moves distinctly toward higher distance values, suggesting that it is able to respond more effectively to the feedback signals provided by the fitness‑sharing mechanism during the LLaMEA loop. 
Overall, these results indicate that the ELA‑based niching strategy successfully promotes diversity in feature space 
and is therefore used in all subsequent experiments.

\section{Results}
\label{sec:results}
In this section, we present the experimental results. 
We analyze them in two ways: 
(1)~through a basin-of-attraction–based methodology providing a more reliable yet computationally expensive verification of each property, and 
(2)~by visualizing the problems in a two-dimensional space using $t$-SNE landscape feature data to confirm that the generated problems fill the gaps in the problem space that are not already represented by the BBOB set.

\begin{figure}[tbp]
  \centering
    \includegraphics[width=0.80\textwidth]{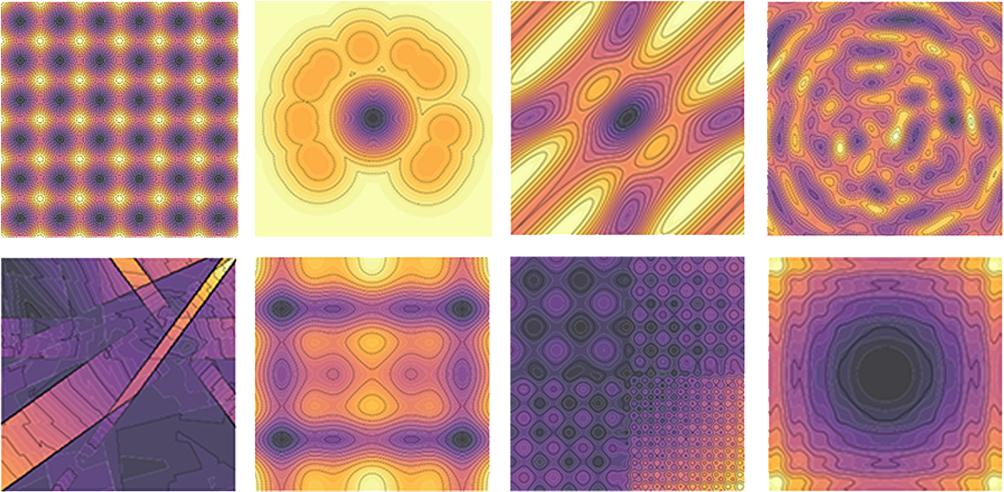}
  \caption{A selection of diverse generated landscapes. With high-level features from left to right, for the top row; Basin-size homogeneity, global to local optima contrast, search space homogeneity, multimodality. For the bottom row: Non-homogenous basin sizes, Non-homogeneous basins with separability, Non-homogenous search space, and separable with homogenous search space. More can be found in the supplementary material.} 
  \label{fig:selection}
\end{figure}

\subsection{Analysis of the High Level Properties}

We provide a detailed analysis of \emph{Multimodality}, \emph{Basin Size Homogeneity}, and \emph{Global–Local Optima Contrast}.  
For each property, we define a metric based on the basin of attraction analysis, and compare it between two classes of problems: (1)~problems specifically optimized for the respective property, and (2)~those that were not. 
\emph{Separability} is excluded since it is mathematically defined, and \emph{Search Space Homogeneity} lacks a clear quantitative metric. 
 However, this property can be assessed through visualizations where non-homogeneous problems are characterized by distinct phase changes in the search space, see Figure~\ref{fig:selection}. 
 It is important to note that even problems not specifically optimized for a specific property might still contain it, especially when considering the large overlap of properties between the BBOB problems. Even so, we expect distinct metric distributions between the two classes.  

We analyse all problems optimized for a given property, including paired-property cases, as each problem is expected to exhibit both target features. We retain only problems where both the combined fitness $\hat F$ and the fitness of the individual property being examined exceed 0.5; the threshold at which the prediction models indicate property presence. 
After filtering, the ratio of the problems that remain are $0.54$ for Multimodality, $0.33$ for Global–Local Optima Contrast, and $0.18$ for Basin Size Homogeneity, with mean problem fitness of $0.97$, $0.91$, and $0.77$ respectively. 

\begin{figure}[tbp]
    \centering
    \begin{subfigure}[t]{0.325\textwidth}
        \centering
        \includegraphics[height=28mm]{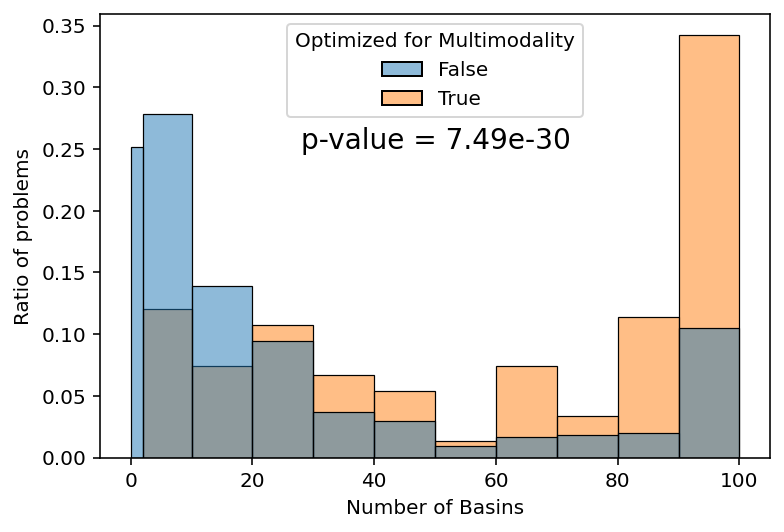}
        \caption{}
        \label{fig:multimodality}
    \end{subfigure}
    \begin{subfigure}[t]{0.325\textwidth}
        \centering
        \includegraphics[height=28mm]{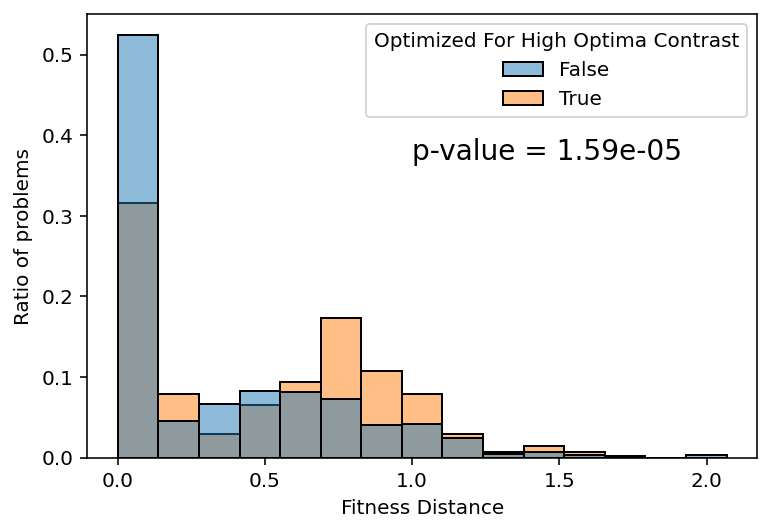}
        \caption{}
        \label{fig:relative_difference}
    \end{subfigure}
    \begin{subfigure}[t]{0.325\textwidth}
        \centering
        \includegraphics[height=28mm]{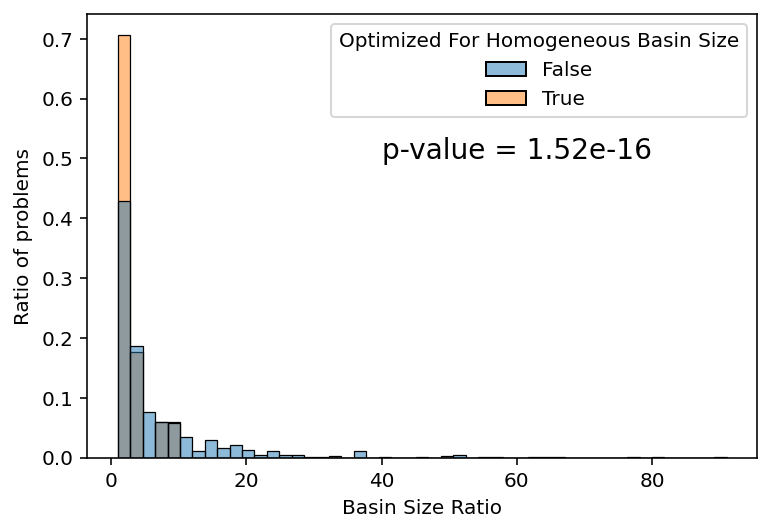}
        \caption{}
        \label{fig:basinsize_results}
    \end{subfigure}
    
    \begin{subfigure}[t]{0.325\textwidth}
        \centering
        \includegraphics[height=28mm]{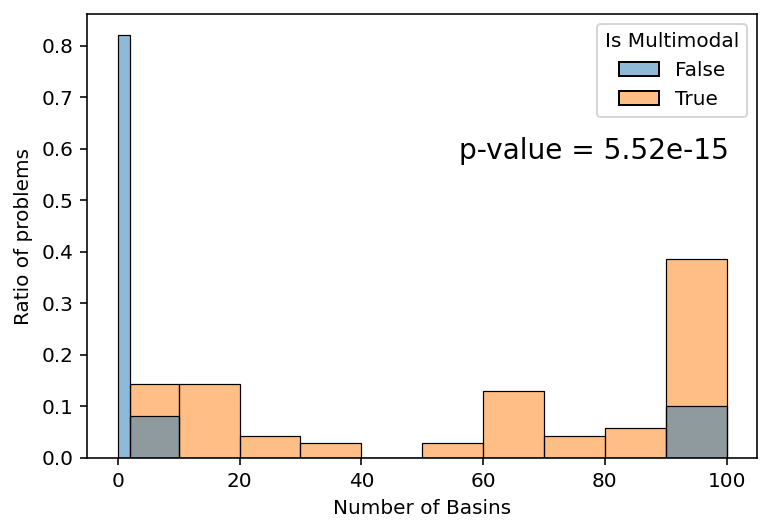}
        \caption{}
        \label{fig:multimodality_bbob}
    \end{subfigure}
    \begin{subfigure}[t]{0.325\textwidth}
        \centering
        \includegraphics[height=28mm]{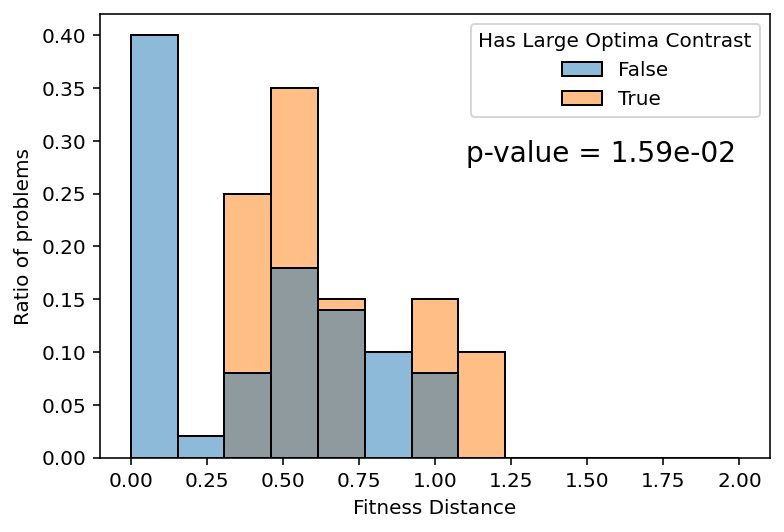}
        \caption{}
        \label{fig:relative_difference_bbob}
    \end{subfigure}
    \begin{subfigure}[t]{0.325\textwidth}
        \centering
        \includegraphics[height=28mm]{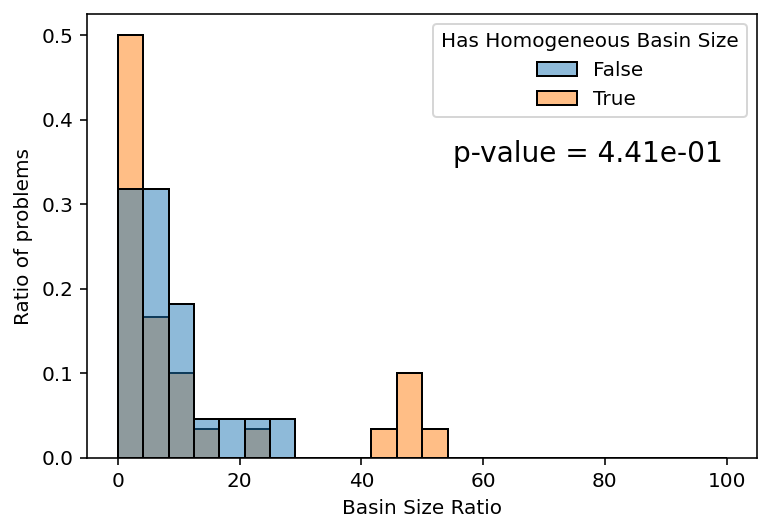}
        \caption{}
        \label{fig:basinsize_results_bbob}
    \end{subfigure}
    \caption{Upper row (a-c): Generated instances optimized for (a)~number of basins of attraction, (b)~relative difference in fitness values between the best found optimum versus the average fitness of all local optima, and (c)~ratio between the number of runs ending in the largest and smallest basin of attraction.\\Lower row (d-f): Same distributions for the BBOB problems. The p-values show the result of a two-tailed Mann-Whitney $U$-test between the distributions. All results except (f) achieve a value lower than $\alpha=0.01$.}
    \label{fig:placeholder}
\end{figure}

Figure~\ref{fig:multimodality} shows the distribution of unique local optima detected by the verification methodology, comparing problems generated by LLaMEA with those from the BBOB set (see Figure~\ref{fig:multimodality_bbob}). The metric performs well for the BBOB problems, with most being unimodal. Two outliers, the Step Ellipsoidal (BBOB Nr.~7) and Bent Cigar (BBOB Nr.~12), are identified as multimodal due to large plateaus that the basin of attraction method interprets as local minima, whereas the original definition does not.
The LLaMEA problems achieve a similar distribution to that of the BBOB problems, and all of the problems that were optimized for multimodality contain more than a single basin of attraction. 

Figure~\ref{fig:relative_difference} (and Figure~\ref{fig:relative_difference_bbob} for BBOB)
shows the absolute difference in fitness values between the single best local optima 
and the average fitness value of all optima. To remove the effects of arbitrary translation and scaling, fitness values are shifted so that the best optimum has a value of 0 and normalized by the mean fitness of the entire search space obtained from 10\,000 uniform samples. Problems with only one optimum are excluded.  

As in previous analyses, some BBOB outliers occur, particularly in rugged functions such as the Rastrigin problem (BBOB Nr.~3). Despite relatively low absolute fitness values, the basin of attraction metric confirms that LLaMEA generates problems with strong global to local optima contrast. 
An additional analysis assessed how the fitness value of the generated problems correlates with this property.
Among problems optimized for high contrast with $F \geq 0.5$, 30\% show small fitness distances; 
filtering for $F \geq 0.9$ reduces this to 25\%, and for $F \geq 0.99$ to 12\%. 
These results verify that higher achieved fitness corresponds to stronger optima contrast.




Figure \ref{fig:basinsize_results} shows the ratio between the smallest and largest basin of attraction. Among the three properties, this one is the most difficult to analyse  with our methodology, as highly multimodal problems can cause every single run of the basin of attraction model to end up in a different unique local optima. Such cases were excluded, since meaningful basin-size estimation is not possible. The number of basins detected also has an effect on their relative size.  When analysing the BBOB data (see Figure~\ref{fig:basinsize_results_bbob}), small differences are visible despite these limitations. The outliers on the right of the plot belong to BBOB Nr.~20, the only problem categorized as having a deceptive global structure, which very strongly guides a large portion of the samples towards a deceptive local optimum. 
For LLaMEA‑generated problems, optimization for homogeneity more often results in smaller ratios between the largest and smallest basins of attraction, indicating a more balanced distribution of basin sizes 





\subsection{Problem Space Visualization}

\begin{figure}[htbp]
  \centering
  \includegraphics[height=27.5mm, trim={35 40 5 5}, clip]{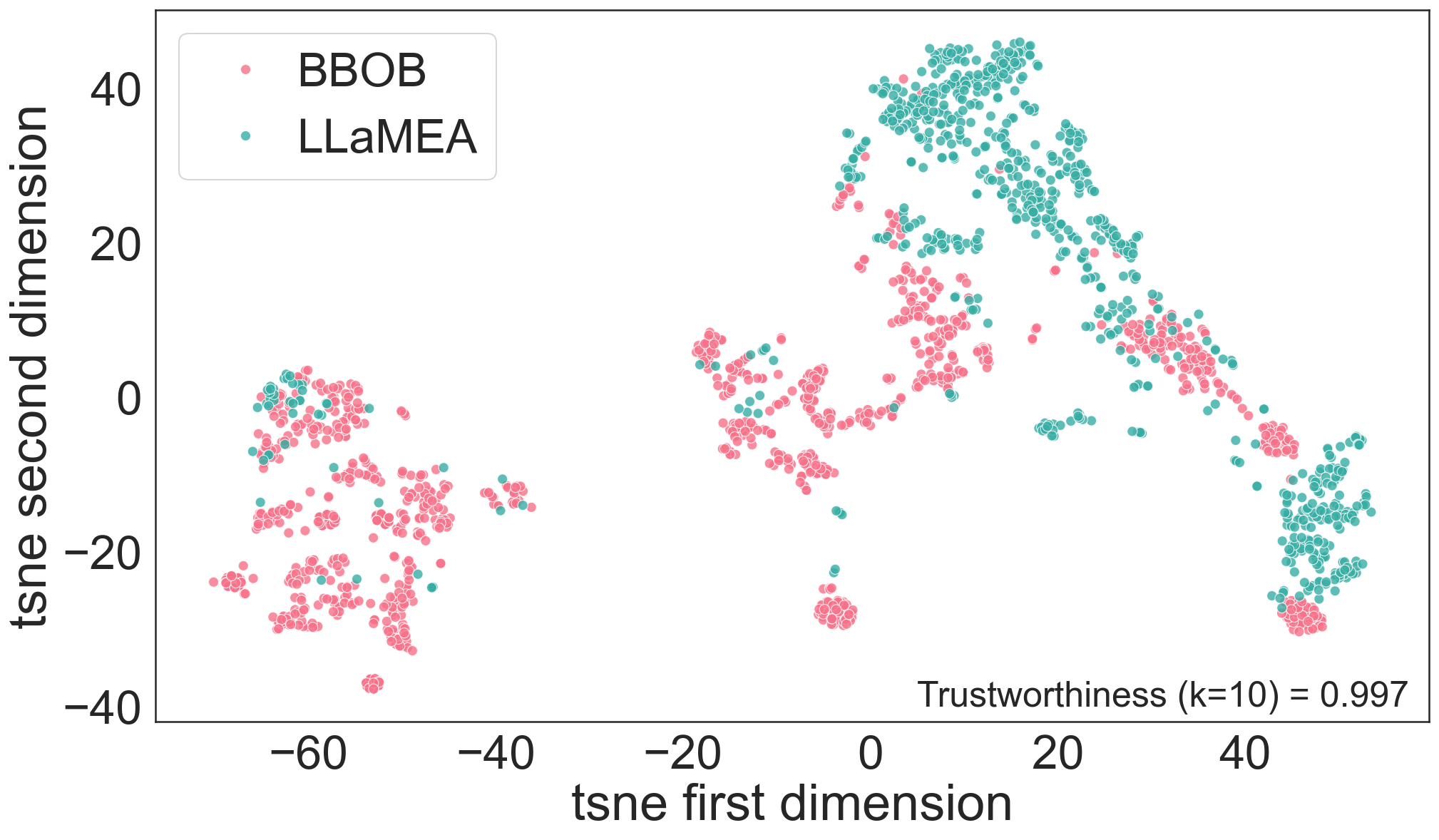}
  \hspace{10mm}
  \includegraphics[height=27.5mm, trim={35 40 5 5}, clip]{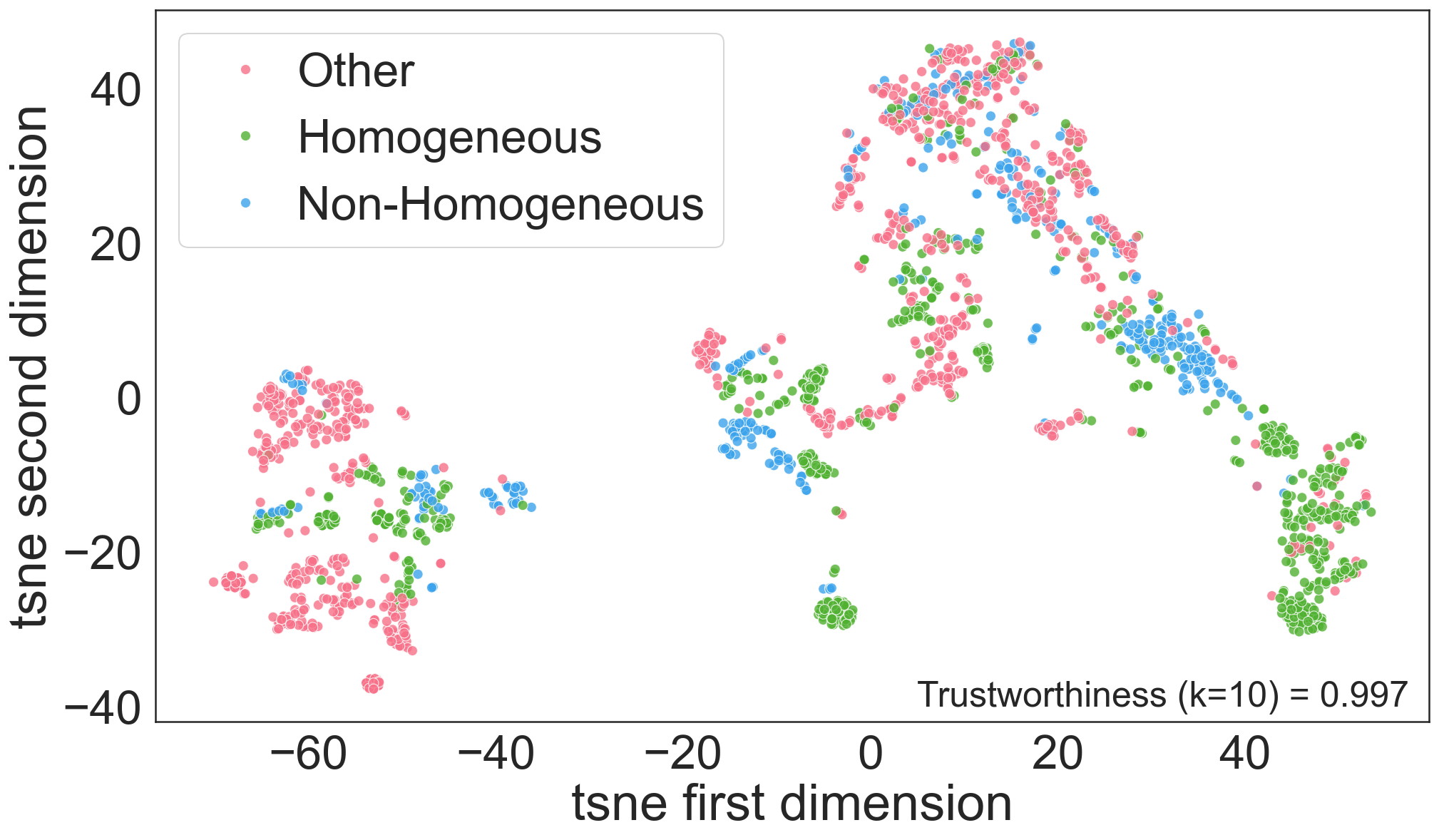}

  \caption{LLaMEA generated problems and BBOB benchmark problems visualized in a 2D feature space. Left: separation by source (BBOB vs. LLaMEA). Right: grouping by basin-size homogeneity.}
  \label{fig:tsne}
\end{figure}

Figure \ref{fig:tsne} shows the visualization of the $t$-SNE \cite{maaten2008visualizing} transformed landscape features of the generated LLM problems, as well as the first 50 instances of all BBOB problems. 
The $t$-SNE transformation allows us to visualize the multivariate landscape feature data in a 2-dimensional space, placing problems with similar features close together. Both $t$-SNE plots have a high trustworthiness scores of nearly 1 (with $k=10$), indicating that local neighbourhoods are well-preserved in the embedded space. The left half of the Figure shows the problems separated by whether they are part of the BBOB benchmark or generated by LLaMEA. We can see that the LLaMEA problems seem to expand the space of the BBOB problems fairly well, without being completely distinct. The right part separates the problems by their properties. However, as there is overlap between the properties, and the LLaMEA problems are trained on pairs of properties, we separate the problems into the following groups:
\begin{description}

\item [Problems with Homogeneous Basin Size]  optimized for homogeneous basin size and not optimized for a large global to local optima contrast.
\item [Problems with non-Homogeneous Basin Size] optimized for global to local optima contrast and additional feature; excluding basin size homogeneity. 
\item[Other Problems] not optimized for multimodality, homogeneous-basin sizes, or global to local optima contrast.
\end{description}
By combining both plots, we observe that most of the BBOB problems categorized as `Other' form a distinct cluster on the left, mainly representing unimodal problems that are comparatively easy and less interesting from a benchmarking perspective. In contrast, the LLaMEA generated problems occupy the remaining regions, expanding the feature space covered by BBOB and thereby increasing the diversity of available benchmark landscapes.

\section{Conclusion}
\label{sec:conclusion}
We demonstrated that Large Language Models, integrated into  an evolutionary loop, can be used to automatically generate continuous optimization problems that exhibit clearly defined high-level landscape properties. By combining LLaMEA's evolutionary prompting with ELA-based property prediction models, we showed that LLMs are able to produce verifiable instances reflecting multimodality, separability, basin-size homogeneity, search-space homogeneity and global–local contrast. Post-generation filtering ensures that only well-fitting instances are retained. We also have shown that the proposed fitness-sharing mechanism effectively increased the diversity of the resulting landscapes in feature space. Complementary basin-of-attraction analyses confirmed structural differences between problem groups, while the $t$-SNE visualisation (Figure \ref{fig:tsne}) indicated that the generated problems expand the BBOB landscape space without forming an unrelated class of their own. 

Overall, our results indicate that LLM-driven problem generation is a promising and flexible direction. The approach offers an interpretable way to target qualitative landscape characteristics and produces a large, openly available library of novel problems that can benefit benchmarking, exploratory landscape studies, and downstream tasks such as automated algorithm selection.
A visual analysis of the problems also shows that the different problem groups contain distinct problems (Figure \ref{fig:selection}), confirming the success in generating problems that are not homogeneous in their search space, and that using combinations of multiple high-level features produces unique problems with properties that are under-represented in the BBOB problem set.

Several opportunities remain for future work. A structured comparison with existing problem generators (including affine combinations, GP-based generation, and other automated methods) is important to position this approach within the broader landscape of problem-design techniques. The sensitivity of the results to the choice and parametrisation of the underlying LLM should also be examined more systematically. Finally, analysing whether the generated instances improve automated algorithm selection or generalise predictive models to unseen problem sets is an important next step. All generated problems and code are publicly available in our repository~\cite{van_stein_2026_18306723}.
\subsubsection{\discintname}
The authors have no competing interests to declare that are relevant to the content of this article.

\bibliographystyle{splncs04}
\bibliography{main}

\end{document}

%% file: figures/abtract_figure.tex
\usetikzlibrary{shapes.misc, positioning}
\begin{tikzpicture}[scale=0.52, transform shape]

\node[] at (-8.5,2.75) () {\bf \Large 1. Definition};
\node[align=center, anchor=north] at (-8.5,-3.0) () {\bf Train predictor Model $M$ \\ \bf with weights $\Theta$};

\node[draw, inner sep=0, line width=0.25, fill=white] at (-11, 1.50) (inst1) {\includegraphics[width=8mm]{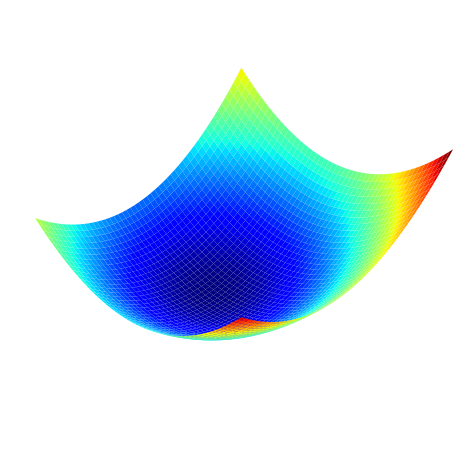}};
\node[draw, inner sep=0, line width=0.25, fill=white] at (-11, 0.50) (inst2) {\includegraphics[width=8mm]{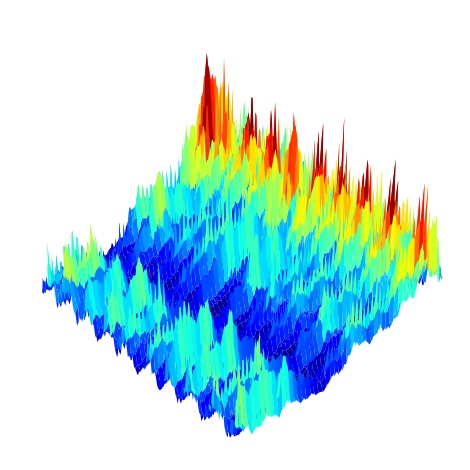}};
\node[draw, inner sep=0, line width=0.25, fill=white] at (-11, -0.50) (inst3) {\includegraphics[width=8mm]{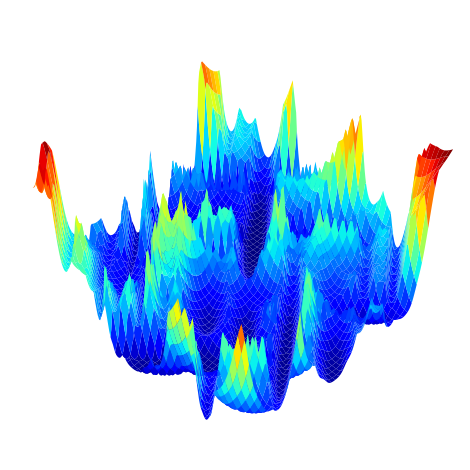}};
\node[draw, inner sep=0, line width=0.25, fill=white] at (-11, -1.50) (inst4) {\includegraphics[width=8mm]{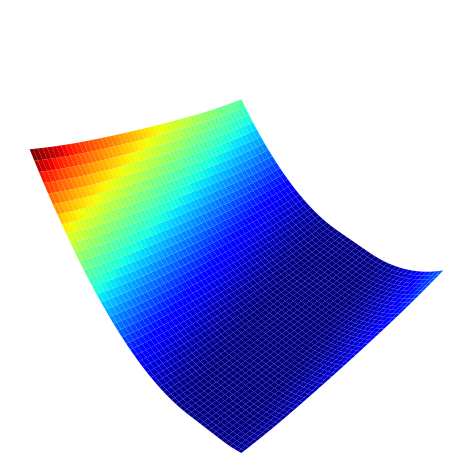}};

\node[] at (-6, 1.50) (label1) {$F_{t=1}\stackrel{!}{=}0$};
\node[] at (-6, 0.50) (label2) {$F_{t=2}\stackrel{!}{=}1$};
\node[] at (-6, -0.50) (label3) {$F_{t=3}\stackrel{!}{=}1$};
\node[] at (-6, -1.50) (label4) {$F_{t=4}\stackrel{!}{=}0$};

\draw[->,dotted] (inst1.east) -- (label1.west);
\draw[->,dotted] (inst2.east) -- (label2.west);
\draw[->,dotted] (inst3.east) -- (label3.west);
\draw[->,dotted] (inst4.east) -- (label4.west);

\node[fill=white, rotate=90, inner sep=5, minimum width=40mm] at (-9.25, 0) () {$\bar{\phi}_{f_t} = \mathbb{E}_{X \sim p_X}\!\left[ \text{ELA}(X, f_t(X)) \right]$};
\node[fill=white, rotate=90, inner sep=5, minimum width=40mm] at (-7.75, 0) () {$F_t = M_\Theta(\bar{\phi_{f_t}})$};

\node[align=center, anchor=north] at (-11, -2.20) () {\tiny Training \\[-5pt] \tiny Instances};
\node[align=center, anchor=north] at (-9.25, -2.20) () {\tiny Compute \\[-5pt] \tiny ELA features};
\node[align=center, anchor=north] at (-7.75, -2.20) () {\tiny Receive Model \\[-5pt] \tiny prediction};
\node[align=center, anchor=north] at (-6, -2.20) () {\it \tiny Is instance \\[-5pt] \it \tiny multimodal?};


\node[] at (0,2.75) () {\bf \Large 2. Generation};
\node[align=center, anchor=north] at (0,-3.0) () {\bf Generate new Instances};
\node[] at (0,0) (max_score) {\LARGE $\max \hat{F}_i$};

\node[draw, rounded corners=1mm, align=center, fill=cyan!40] at (3.0,0) (gen_new_instance) {Receive generated\\Instance $f_i \in \mathcal{I}$};
\node[draw, rounded corners=1mm, align=center, fill=cyan!70] at (0.,1.50) (compute_new_features) {Compute ELA features using\\$\bar{\phi}_{f_i} = \mathbb{E}_{X \sim p_X}\!\left[ \text{ELA}(X, f_i(X)) \right]$};
\node[draw, rounded corners=1mm, align=center, fill=cyan] at (-3.0,0) (compute_new_score) {Get fittness\\$F_i = M_\Theta(\bar{\phi}_{f_i})$};
\node[draw, rounded corners=1mm, align=center, fill=BurntOrange!80] at (0,-1.50) (llamea) {Provide $\hat{F}_i$ to LLaMEA to\\further improve model score};

\draw[->, line width=1.0pt] (gen_new_instance.north) to[out=90,in=0] (compute_new_features.east);
\draw[->, line width=1.0pt] (compute_new_features.west) to[out=180,in=90] (compute_new_score.north);
\draw[->, line width=1.0pt] (compute_new_score.south) to[out=270,in=180] (llamea.west);
\draw[->, line width=1.0pt] (llamea.east) to[out=0,in=270] (gen_new_instance.south);

\node[align=center, anchor=north] at (0, -2.40) () {\it \tiny Repeat until temination criteria is met!};

\node[] at (8.50,2.75) () {\bf \Large 3. Verification};
\node[align=center, anchor=north] at (8.50,-3.0) () {\bf Filter and verify new instances};

\node[draw, inner sep=0, line width=0.25, fill=white] at (5.75, 1.50) (new_inst1) {\includegraphics[width=8mm]{figures/basins-example.png}};
\node[draw, inner sep=0, line width=0.25, fill=white] at (5.75, 0.50) (new_inst2) {\includegraphics[width=8mm]{figures/basins-example.png}};
\node[draw, inner sep=0, line width=0.25, fill=white] at (5.75, -0.50) (new_inst3) {\includegraphics[width=8mm]{figures/landscape-example.png}};
\node[draw, inner sep=0, line width=0.25, fill=white] at (5.75, -1.50) (new_inst4) {\includegraphics[width=8mm]{figures/landscape-example.png}};

\node[align=right, anchor=east] at (11.75, 1.50) (count_basins) {Count\\Basins};
\node[align=right, anchor=east] at (11.75, 0.50) (basins_size) {Basins\\Size};
\node[align=right, anchor=east] at (11.75, -0.50) (optima_contrast) {Optima\\Contrast};
\node[align=right, anchor=east] at (11.75, -1.50) (space_homogeneity) {Search Space\\Homogeneity};

\draw[->,dotted] (new_inst1.east) -- (count_basins.west);
\draw[->,dotted] (new_inst2.east) -- (basins_size.west);
\draw[->,dotted] (new_inst3.east) -- (optima_contrast.west);
\draw[->,dotted] (new_inst4.east) -- (space_homogeneity.west);

\node[align=center, anchor=north] at (8.50, -2.20) () {\it \tiny Remove generated instances \\[-5pt] \it \tiny that do not fulfill requirements!};

\node[fill=white, rotate=90, inner sep=5, minimum width=40mm] at (8.0, 0) () {Verify Instance to meet};

\draw[dashed] (-4.75,2.25) -- (-4.75,-3.5);
\draw[dashed] (4.75,2.25) -- (4.75,-3.5);
\end{tikzpicture}

%% file: main.bib
@inproceedings{antonov2023new,
  title={New solutions to Cooke triplet problem via analysis of attraction basins},
  author={Antonov, Kirill and Botari, Tiago and Tukker, Teuss and B{\"a}ck, Thomas and van Stein, Niki and Kononova, Anna V},
  booktitle={Digital Optical Technologies 2023},
  volume={12624},
  pages={131--143},
  year={2023},
  organization={SPIE}
}

@ARTICLE{van2025llamea,
  author={Stein, Niki van and Bäck, Thomas},
  journal={IEEE Transactions on Evolutionary Computation},
  title={{LLaMEA}: A Large Language Model Evolutionary Algorithm for Automatically Generating Metaheuristics},
  year={2025},
  volume={29},
  number={2},
  pages={331-345},
  doi={10.1109/TEVC.2024.3497793}
}

@article{van2025behaviour,
  title={Behaviour space analysis of llm-driven meta-heuristic discovery},
  author={van Stein, Niki and Yin, Haoran and Kononova, Anna V and B{\"a}ck, Thomas and Ochoa, Gabriela},
  journal={arXiv preprint arXiv:2507.03605},
  year={2025}
}

@inproceedings{10.1145/3712255.3734288,
author = {Yin, Haoran and Kononova, Anna V. and B\"{a}ck, Thomas and van Stein, Niki},
title = {Optimizing Photonic Structures with Large Language Model Driven Algorithm Discovery},
year = {2025},
isbn = {9798400714641},
publisher = {Association for Computing Machinery},
address = {New York, NY, USA},
url = {https://doi.org/10.1145/3712255.3734288},
doi = {10.1145/3712255.3734288},
booktitle = {Proceedings of the Genetic and Evolutionary Computation Conference Companion},
pages = {2354–2362},
numpages = {9},
location = {NH Malaga Hotel, Malaga, Spain},
series = {GECCO '25 Companion}
}

@inproceedings{liu2024evolution,
  title={Evolution of heuristics: towards efficient automatic algorithm design using large language model},
  author={Liu, Fei and Tong, Xialiang and Yuan, Mingxuan and Lin, Xi and Luo, Fu and Wang, Zhenkun and Lu, Zhichao and Zhang, Qingfu},
  booktitle={Proceedings of the 41st International Conference on Machine Learning},
  pages={32201--32223},
  year={2024}
}

@article{ye2024reevo,
  title={Reevo: Large language models as hyper-heuristics with reflective evolution},
  author={Ye, Haoran and Wang, Jiarui and Cao, Zhiguang and Berto, Federico and Hua, Chuanbo and Kim, Haeyeon and Park, Jinkyoo and Song, Guojie},
  journal={Advances in neural information processing systems},
  volume={37},
  pages={43571--43608},
  year={2024}
}

@article{zheng2025monte,
  title={Monte carlo tree search for comprehensive exploration in llm-based automatic heuristic design},
  author={Zheng, Zhi and Xie, Zhuoliang and Wang, Zhenkun and Hooi, Bryan},
  journal={arXiv preprint arXiv:2501.08603},
  year={2025}
}

@book{10.5555/548530,
author = {B{\"a}ck, Thomas and Fogel, David B. and Michalewicz, Zbigniew},
title = {Handbook of Evolutionary Computation},
year = {1997},
isbn = {0750303921},
publisher = {IOP Publishing Ltd.},
address = {GBR},
edition = {1st}
}

@article{van_Stein_2025,
   title={In-the-loop Hyper-Parameter Optimization for LLM-Based Automated Design of Heuristics},
   ISSN={2688-3007},
   url={http://dx.doi.org/10.1145/3731567},
   DOI={10.1145/3731567},
   journal={ACM Transactions on Evolutionary Learning and Optimization},
   publisher={Association for Computing Machinery (ACM)},
   author={van Stein, Niki and Vermetten, Diederick and Bäck, Thomas},
   year={2025},
   month=apr }

@inproceedings{10.1145/3205651.3208284,
author = {Ullrich, Markus and Weise, Thomas and Awasthi, Abhishek and L\"{a}ssig, J\"{o}rg},
title = {A generic problem instance generator for discrete optimization problems},
year = {2018},
isbn = {9781450357647},
publisher = {Association for Computing Machinery},
address = {New York, NY, USA},
url = {https://doi.org/10.1145/3205651.3208284},
doi = {10.1145/3205651.3208284},
booktitle = {Proceedings of the Genetic and Evolutionary Computation Conference Companion},
pages = {1761–1768},
numpages = {8},
location = {Kyoto, Japan},
series = {GECCO '18}
}

@ARTICLE{9185169,
  author={Muñoz, Mario A. and Smith-Miles, Kate},
  journal={Evolutionary Computation}, 
  title={Generating New Space-Filling Test Instances for Continuous Black-Box Optimization}, 
  year={2020},
  volume={28},
  number={3},
  pages={379-404},
  doi={10.1162/evco_a_00262}}

@ARTICLE{9187549,
  author={Tian, Ye and Peng, Shichen and Zhang, Xingyi and Rodemann, Tobias and Tan, Kay Chen and Jin, Yaochu},
  journal={IEEE Transactions on Artificial Intelligence}, 
  title={A Recommender System for Metaheuristic Algorithms for Continuous Optimization Based on Deep Recurrent Neural Networks}, 
  year={2020},
  volume={1},
  number={1},
  pages={5-18},
  doi={10.1109/TAI.2020.3022339}}

@misc{wang2025instancegenerationmetablackboxoptimization,
      title={Instance Generation for Meta-Black-Box Optimization through Latent Space Reverse Engineering}, 
      author={Chen Wang and Zeyuan Ma and Zhiguang Cao and Yue-Jiao Gong},
      year={2025},
      eprint={2509.15810},
      archivePrefix={arXiv},
      primaryClass={cs.LG},
      url={https://arxiv.org/abs/2509.15810}, 
}

@inproceedings{10.1145/3712255.3726663,
author = {Ono, Yuhiro and Harada, Tomohiro and Miura, Yukiya},
title = {Large Language Model Driven Evolutionary Optimization Benchmark Generation Algorithm},
year = {2025},
isbn = {9798400714641},
publisher = {Association for Computing Machinery},
address = {New York, NY, USA},
url = {https://doi.org/10.1145/3712255.3726663},
doi = {10.1145/3712255.3726663},
booktitle = {Proceedings of the Genetic and Evolutionary Computation Conference Companion},
pages = {151–154},
numpages = {4},
location = {NH Malaga Hotel, Malaga, Spain},
series = {GECCO '25 Companion}
}

@inproceedings{mersmann2010benchmarking,
  title={Benchmarking evolutionary algorithms: Towards exploratory landscape analysis},
  author={Mersmann, Olaf and Preuss, Mike and Trautmann, Heike},
  booktitle={International Conference on Parallel Problem Solving from Nature},
  pages={73--82},
  year={2010},
  organization={Springer}
}

@article{pflacco,
    author = {Prager, Raphael Patrick and Trautmann, Heike},
    title = "{Pflacco: Feature-Based Landscape Analysis of Continuous and Constrained Optimization Problems in Python}",
    journal = {Evolutionary Computation},
    pages = {1-25},
    year = {2023},
    month = {07},
    issn = {1063-6560},
    doi = {10.1162/evco_a_00341},
    url = {https://doi.org/10.1162/evco\_a\_00341},
    eprint = {https://direct.mit.edu/evco/article-pdf/doi/10.1162/evco\_a\_00341/2148122/evco\_a\_00341.pdf},
}

@article{mckay2000comparison,
  title={A comparison of three methods for selecting values of input variables in the analysis of output from a computer code},
  author={McKay, Michael D and Beckman, Richard J and Conover, William J},
  journal={Technometrics},
  volume={42},
  number={1},
  pages={55--61},
  year={2000},
  publisher={Taylor \& Francis}
}

@inproceedings{seiler2022collection,
  title={A collection of deep learning-based feature-free approaches for characterizing single-objective continuous fitness landscapes},
  author={Seiler, Moritz Vinzent and Prager, Raphael Patrick and Kerschke, Pascal and Trautmann, Heike},
  booktitle={Proceedings of the Genetic and Evolutionary Computation Conference},
  pages={657--665},
  year={2022}
}

@inproceedings{mersmann2011exploratory,
  title={Exploratory landscape analysis},
  author={Mersmann, Olaf and Bischl, Bernd and Trautmann, Heike and Preuss, Mike and Weihs, Claus and Rudolph, G{\"u}nter},
  booktitle={Proceedings of the 13th annual conference on Genetic and evolutionary computation},
  pages={829--836},
  year={2011}
}

@inproceedings{chen2016xgboost,
  title={Xgboost: A scalable tree boosting system},
  author={Chen, Tianqi and Guestrin, Carlos},
  booktitle={Proceedings of the 22nd acm sigkdd international conference on knowledge discovery and data mining},
  pages={785--794},
  year={2016}
}

@article{maaten2008visualizing,
  title={Visualizing data using t-SNE},
  author={Maaten, Laurens van der and Hinton, Geoffrey},
  journal={Journal of machine learning research},
  volume={9},
  number={Nov},
  pages={2579--2605},
  year={2008}
}

@article{cenikj2025landscape,
  title={Landscape features in single-objective continuous optimization: Have we hit a wall in algorithm selection generalization?},
  author={Cenikj, Gjorgjina and Petelin, Ga{\v{s}}per and Seiler, Moritz and Cenikj, Nikola and Eftimov, Tome},
  journal={Swarm and Evolutionary Computation},
  volume={94},
  pages={101894},
  year={2025},
  publisher={Elsevier}
}

@inproceedings{seiler2025randoptgen,
  title={RandOptGen: A Unified Random Problem Generator for Single-and Multi-Objective Optimization Problems with Mixed-Variable Input Spaces},
  author={Seiler, Moritz Vinzent and Preu{\ss}, Oliver Ludger and Trautmann, Heike},
  booktitle={Proceedings of the Genetic and Evolutionary Computation Conference},
  pages={76--84},
  year={2025}
}

@inproceedings{vermetten2023ma,
  title={{MA-BBOB}: Many-affine combinations of bbob functions for evaluating automl approaches in noiseless numerical black-box optimization contexts},
  author={Vermetten, Diederick and Ye, Furong and B{\"a}ck, Thomas and Doerr, Carola},
  booktitle={International Conference on Automated Machine Learning},
  pages={7--1},
  year={2023},
  organization={PMLR}
}

@article{vermetten2025ma,
  title={{MA-BBOB}: A problem generator for black-box optimization using affine combinations and shifts},
  author={Vermetten, Diederick and Ye, Furong and B{\"a}ck, Thomas and Doerr, Carola},
  journal={ACM Transactions on Evolutionary Learning and Optimization},
  volume={5},
  number={1},
  pages={1--19},
  year={2025},
  publisher={ACM New York, NY}
}

@inproceedings{dietrich2022increasing,
  title={Increasing the diversity of benchmark function sets through affine recombination},
  author={Dietrich, Konstantin and Mersmann, Olaf},
  booktitle={International Conference on Parallel Problem Solving from Nature},
  pages={590--602},
  year={2022},
  organization={Springer}
}

@article{tian2020recommender,
 author = {Tian, Ye and Peng, Shichen and Zhang, Xingyi and Rodemann, Tobias and Tan, Kay Chen and Jin, Yaochu},
 journal = {{IEEE Transactions on Artificial Intelligence}},
 number = {1},
 pages = {5--18},
 publisher = {IEEE},
 title = {{A Recommender System for Metaheuristic Algorithms for Continuous Optimization Based on Deep Recurrent Neural Networks}},
 volume = {1},
 year = {2020}
}

@article{kerschke2019AS,
  author={Kerschke, Pascal and Hoos, Holger H. and Neumann, Frank and Trautmann, Heike},
  journal={Evolutionary Computation}, 
  title={Automated Algorithm Selection: Survey and Perspectives}, 
  year={2019},
  volume={27},
  number={1},
  pages={3-45},
  doi={10.1162/evco_a_00242}}

@techreport{hansen:inria-00362633,
  TITLE = {{Real-Parameter Black-Box Optimization Benchmarking 2009: Noiseless Functions Definitions}},
  AUTHOR = {Hansen, Nikolaus and Finck, Steffen and Ros, Raymond and Auger, Anne},
  URL = {https://inria.hal.science/inria-00362633},
  TYPE = {Research Report},
  NUMBER = {RR-6829},
  INSTITUTION = {{INRIA}},
  YEAR = {2009},
  HAL_ID = {inria-00362633},
  HAL_VERSION = {v2},
}

@misc{fan2024surveyragmeetingllms,
      title={A Survey on RAG Meeting LLMs: Towards Retrieval-Augmented Large Language Models}, 
      author={Wenqi Fan and Yujuan Ding and Liangbo Ning and Shijie Wang and Hengyun Li and Dawei Yin and Tat-Seng Chua and Qing Li},
      year={2024},
      eprint={2405.06211},
      archivePrefix={arXiv},
      primaryClass={cs.CL},
      url={https://arxiv.org/abs/2405.06211}, 
}

@techreport{bbob2019,
    author = {Finck, Steffen and Hansen, Nikolaus and Ros, Raymond and Auger, Anne},
    title = {Real-Parameter Black-Box Optimization Benchmarking 2009: Noiseless Functions Definitions},
    institution = {INRIA},
    year = {2009},
    number = {RR-6829},
    note = {Updated version as of February 2019},
    url = {https://inria.hal.science/inria-00362633v2/document}
}

@software{van_stein_2026_18306723,
  author       = {van Stein, Niki and
                  Škvorc, Urban and
                 Seiler, Moritz and Grimme, Britta and B\"{a}ck, Thomas and Trautmann, Heike},
  title        = {Reproducibility package for "LLM Driven Design of
                   Continuous Optimization Problems".
                  },
  month        = jan,
  year         = 2026,
  publisher    = {Zenodo},
  version      = {paper-ela},
  doi          = {10.5281/zenodo.18306723},
  url          = {https://doi.org/10.5281/zenodo.18306723},
}
